\documentclass[10pt,twocolumn,letterpaper]{article}

\pdfoutput=1
\usepackage{cvpr}
\usepackage{times}
\usepackage{graphicx}
\usepackage{amsmath}
\usepackage{amssymb}
\usepackage{array}
\usepackage{xcolor}

\newcommand{\modelname}{EmotiCon}
\newcommand{\dataname}{GroupWalk}
\usepackage{multirow, makecell}
\usepackage{subcaption}
\usepackage{booktabs}
\usepackage{amsmath}
\usepackage{amssymb}
\usepackage{amsfonts}
\usepackage{enumitem}
\usepackage{float}
\usepackage{bbm}
\usepackage{dsfont}
\usepackage[font=small,belowskip=-1.5pt]{caption} 
\usepackage{bbold} 
\usepackage{hhline}

\newcommand{\shorteq}{%
  \settowidth{\@tempdima}{-}
  \resizebox{\@tempdima}{\height}{=}%
}

\newcolumntype{L}[1]{>{\raggedright\let\newline\\\arraybackslash\hspace{0pt}}m{#1}}
\newcolumntype{C}[1]{>{\centering\let\newline\\\arraybackslash\hspace{0pt}}m{#1}}
\newcolumntype{R}[1]{>{\raggedleft\let\newline\\\arraybackslash\hspace{0pt}}m{#1}}

\usepackage[colorlinks,bookmarks=false]{hyperref}
\cvprfinalcopy 

\begin{document}

\title{\modelname: Context-Aware Multimodal Emotion Recognition using Frege's Principle}
\author{Trisha Mittal\textsuperscript{\rm 1}, Pooja Guhan\textsuperscript{\rm 1}, Uttaran Bhattacharya\textsuperscript{\rm 1}, Rohan Chandra\textsuperscript{\rm 1}, {Aniket Bera\textsuperscript{\rm 1}, Dinesh Manocha\textsuperscript{\rm 1,\rm 2}}\\ 
\textsuperscript{\rm 1}Department of Computer Science, University of Maryland, College Park, USA\\ 
\textsuperscript{\rm 2}Department of Electrical and Computer Engineering, University of Maryland, College Park, USA\\ 
\{trisha, pguhan, uttaranb, rohan, ab, dm\}@cs.umd.edu\\ 
Project URL: \url{https://gamma.umd.edu/emoticon}
}

\maketitle

\begin{abstract}
   We present \modelname, a learning-based algorithm for context-aware perceived human emotion recognition from videos and images. Motivated by Frege's Context Principle from psychology, our approach combines three interpretations of context for emotion recognition. Our first interpretation is based on using multiple modalities~(e.g. faces and gaits) for emotion recognition. For the second interpretation, we gather semantic context from the input image and use a self-attention-based CNN to encode this information. Finally, we use depth maps to model the third interpretation related to socio-dynamic interactions and proximity among agents. We demonstrate the efficiency of our network through experiments on EMOTIC, a benchmark dataset. We report an Average Precision (AP) score of $35.48$ across $26$ classes, which is an improvement of $7$-$8$ over prior methods. We also introduce a new dataset, \dataname, which is a collection of videos captured in multiple real-world settings of people walking. We report an AP of $65.83$ across $4$ categories on \dataname, which is also an improvement over prior methods. 
\end{abstract}
\section{Introduction}
\label{sec:intro}
Perceiving the emotions of people around us is vital in everyday life. Humans often alter their behavior while interacting with others based on their perceived emotions. In particular, automatic emotion recognition has been used for different applications, including human-computer interaction~\cite{ER-hci}, surveillance~\cite{ER-surveillance}, robotics, games, entertainment, and more. Emotions are modeled as either discrete categories or as points in a continuous space of affective dimensions~\cite{pad}. In the continuous space, emotions are treated as points in a 3D space of valence, arousal, and dominance. In this work, our focus is on recognizing perceived human emotion rather than the actual emotional state of a person in the discrete emotion space.

\begin{figure}[t]
    \centering
    \includegraphics[width=\columnwidth]{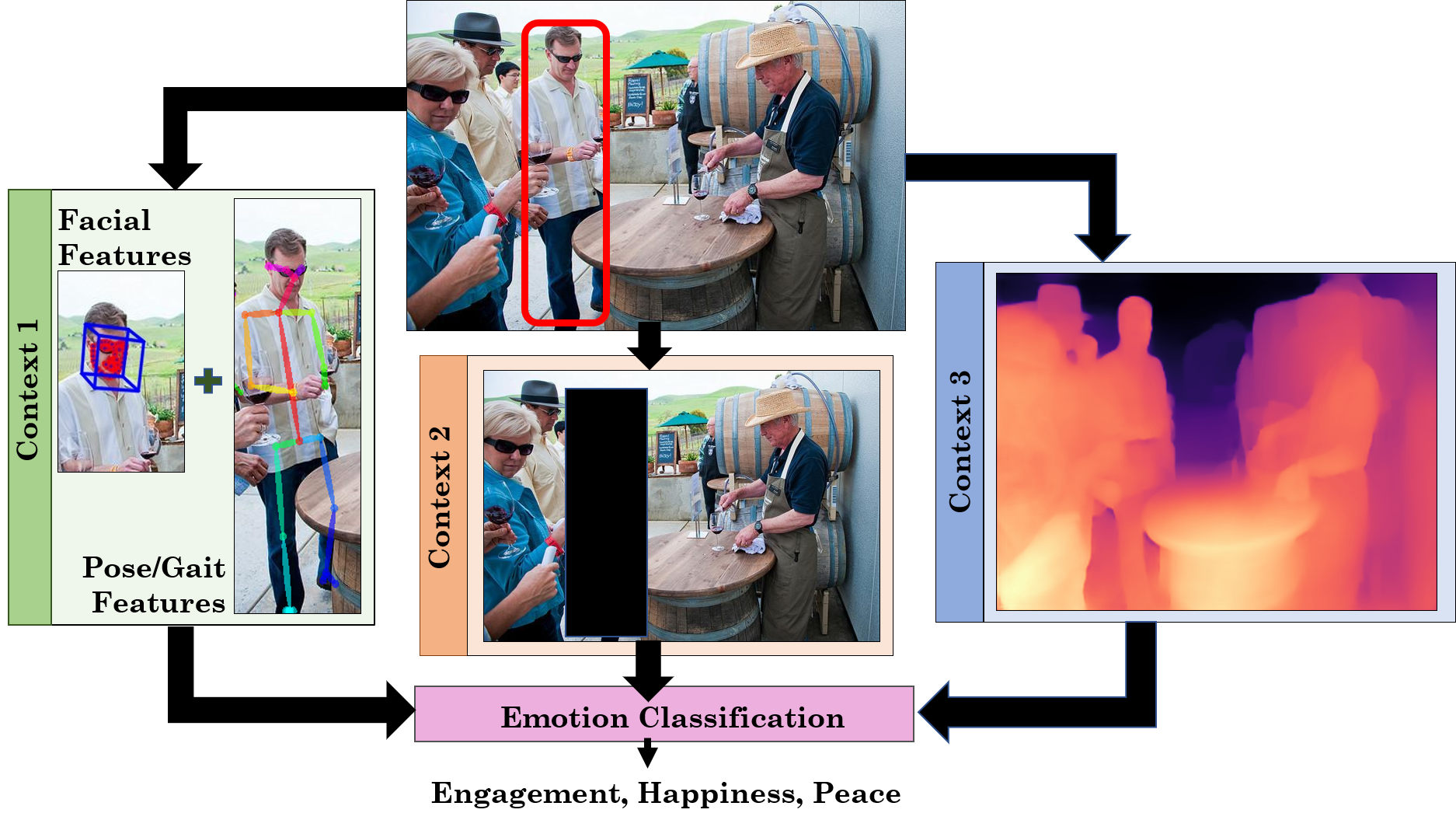}
    \caption{\small{\textbf{Context-Aware Multimodal Emotion Recognition: }We use three interpretations of context to perform perceived emotion recognition. We use multiple modalities~(Context 1) of faces and gaits, background visual information~(Context 2) and socio-dynamic inter-agent interactions~(Context 3) to infer the perceived emotion. \modelname~outperforms prior context-aware emotion recognition methods. Above is an input sample from the EMOTIC dataset.}}
    \label{fig:cover}
\vspace{-10pt}
\end{figure}
 
Initial works in emotion recognition have been mostly unimodal~\cite{face1,face2,speech1,gait} approaches. The unique modality may correspond to facial expressions, voice, text, body posture, gaits, or physiological signals. This was followed by multimodal emotion recognition~\cite{early-fusion, late-fusion, hybrid-fusion}, where various combinations of modalities were used and combined in various manners to infer emotions. 

Although such modalities or cues extracted from a person can provide us with information regarding the perceived emotion, context also plays a very crucial role in the understanding of the perceived emotion. Frege's context principle~\cite{frege}~urges not asking for the meaning of a word in isolation and instead of finding the meaning in the context of a sentence. We use this notion behind the context principle in psychology for emotion recognition. `Context' has been interpreted in multiple ways by researchers in psychology, including:
\begin{enumerate}[label=(\alph*), noitemsep]
    \item \textit{Context 1 (Multiple Modalities): } Incorporating cues from different modalities was one of the initial definitions of context. This domain is also known as Multimodal Emotion Recognition. Combining modalities provides complementary information, which leads to better inference and also performs better on in-the-wild datasets.
    \item \textit{Context 2 (Background Context): } Semantic understanding of the scene from visual cues in the image helps in getting insights about the agent's surroundings and activity, both of which can affect the perceived emotional state of the agent. 
    \item \textit{Context 3 (Socio-Dynamic Inter-Agent Interactions): }Researchers in psychology suggest that the presence or absence of other agents affects the perceived emotional state of an agent. When other agents share an identity or are known to the agent, they often coordinate their behaviors. This varies when other agents are strangers. Such interactions and proximity to other agents have been less explored for perceived emotion recognition.
\end{enumerate}
One of our goals is to make Emotion Recognition systems work for real-life scenarios. This implies using modalities that do not require sophisticated equipment to be captured and are readily available. Psychology researchers~\cite{facegait2} have conducted experiments by mixing faces and body features corresponding to different emotions and found that participants guessed the emotions that matched the body features. This is also because of the ease of ``mocking'' one's facial expressions. Subsequently, researchers~\cite{facegait1,facegait3} found the combination of faces and body features to be a reliable measure of inferring human emotion. As a result, it would be useful to combine such face and body features for context-based emotion recognition.

\paragraph{Main Contributions: } We propose \modelname, a context-aware emotion recognition model. The input to~\modelname~is images/video frames, and the output is a multi-label emotion classification. The novel components of our work include:
\begin{enumerate}[noitemsep]
    \item We present a context-aware multimodal emotion recognition algorithm called \modelname. Consistent with Ferge's Context principle, in this work, we try to incorporate three interpretations of context to perform emotion recognition from videos and images.
    \item We also present a new approach to modeling the socio-dynamic interactions between agents using a depth-based CNN. We compute a depth map of the image and feed that to the network to learn about the proximity of agents to each other.
    \item Though extendable to any number of modalities, we release a new dataset \dataname~for emotion recognition. To the best of our knowledge, there exist very few datasets captured in uncontrolled settings with both faces and gaits that have emotion label annotations. To enable research in this domain, we make ~\dataname~publicly available with emotion annotations. \dataname~is a collection of 45 videos captured in multiple real-world settings of people walking in dense crowd settings. The videos have about $3544$ agents annotated with their emotion labels. 
\end{enumerate}

We compare our work with prior methods by testing our performance on EMOTIC~\cite{emotic}, a benchmark dataset for context-aware emotion recognition. We report an improved AP score of $35.48$ on EMOTIC, which is an improvement of $7-8$ over prior methods~\cite{context1,context2,context3}. We also report AP scores of our approach and prior methods on the new dataset,~\dataname. We perform ablation experiments on both datasets, to justify the need for the three components of \modelname. As per the annotations provided in EMOTIC, we perform a multi-label classification over $26$ discrete emotion labels. On~\dataname~too, we perform a multi-label classification over $4$ discrete emotions~({anger, happy, neutral, sad}). 
\section{Related Work}
\label{sec:relatedwork}
In this section, we give a brief overview of previous works on unimodal and multimodal emotion recognition, context-aware emotion recognition, and existing context-aware datasets. 
\begin{figure*}[t]
    \centering
    \includegraphics[width=\textwidth]{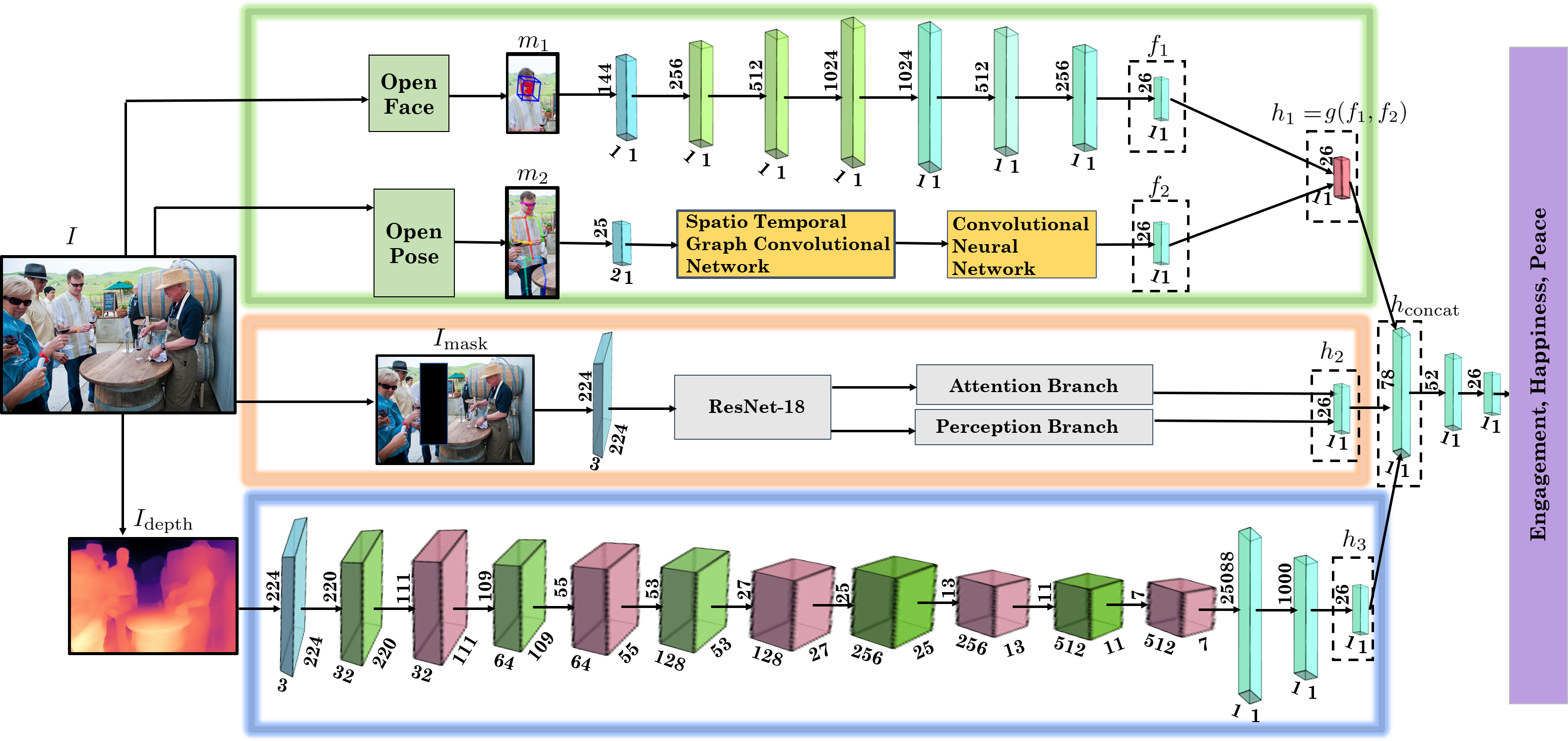}
    \caption{\small{\textbf{\modelname: }We use three interpretations of context. We first extract features for the two modalities to obtain $f_{1}$ and $f_{2}$ and inputs $I_\textrm{mask}$ and $I_\textrm{depth}$ from the raw input image, $I$. These are then passed through the respective neural networks to obtain $h_1$, $h_2$ and $h_3$. To obtain $h_1$, we use a multiplicative fusion layer~(red color) to fuse inputs from both modalities, faces, and gaits.  $h_1$, $h_2$ and $h_3$ are then concatenated to obtain $h_\textrm{concat}$. }}
    \label{fig:overview}
\end{figure*}
\subsection{Uni/Multimodal Emotion Recognition}
Prior works in emotion recognition from handcrafted features~\cite{handcrafted1,handcrafted2} or deep learning networks~\cite{dl1,dl2,dl3} have used single modalities like facial expressions~\cite{face1,face2}, voice, and speech expressions~\cite{speech1}, body gestures~\cite{body}, gaits~\cite{gait}, and physiological signals such as respiratory and heart cues~\cite{physiological1}. There has been a shift in the paradigm, where researchers have tried to fuse multiple modalities to perform emotion recognition, also known as Multimodal Emotion Recognition. Fusion methods like early fusion~\cite{early-fusion}, late fusion~\cite{late-fusion}, and hybrid fusion~\cite{hybrid-fusion} have been explored for emotion recognition from multiple modalities. Multimodal emotion recognition has been motivated by research in psychology and also helped in improving accuracy on in-the-wild emotion recognition datasets like IEMOCAP~\cite{iemocap}  and CMU-MOSEI~\cite{cmu-mosei}. 
\subsection{Context-Aware Emotion Recognition in Psychology Research}
Though introduced in the domain of philosophy of language, Frege~\cite{frege}~proposed that words should never be seen in isolation but in the context of their proposition. Researchers in psychology~\cite{psych1,psych2,psych3} also agree that just like most psychological processes, emotional processes cannot be interpreted without context. They suggest that context often produces emotion and also shapes how emotion is perceived. Emotion literature that addresses context~\cite{psych4,psych5,psych6} suggests several broad categories of contextual features: person, situation, and context. Martinez et al.~\cite{psych8} conduct experiments about the necessity of context and found that even when the participants' faces and bodies were masked in silent videos, viewers were able to infer the affect successfully. Greenway et al.~\cite{psych7} organize these contextual features in three levels, ranging from micro-level~(person) to macro-level~(cultural). In level 2~(situational), they include factors like the presence and closeness of other agents. Research shows that the simple presence of another person elicits more expression of emotion than situations where people are alone~\cite{psych9,psych10}. These expressions are more amplified when people know each other and are not strangers~\cite{psych10}. 
\subsection{Context-Aware Emotion Recognition}
Recent works in context-aware emotion recognition are based on deep-learning network architectures. Kosti et al.~\cite{context1} and Lee et al.~\cite{context2} present two recent advances in context-aware emotion recognition and they propose similar architectures. Both of them have two-stream architectures followed by a fusion network. One stream focuses on a modality~(face for~\cite{context2} and body for~\cite{context1}) and the other focuses on capturing context. Lee et al.~\cite{context2} consider everything other than the face as context, and hence mask the face from the image to feed to the context stream. On the other hand,~\cite{context2} uses a Region Proposal Network~(RPN) to extract context elements from the image. These elements become the nodes of an affective graph, which is fed into a Graph Convolution Network~(GCN) to encode context. 
Another problem that has been looked into is group emotion recognition~\cite{group1,group2}. The objective here is to label the emotion of the entire set of people in the frame under the assumption that they all share some social identity.
\subsection{Context-Aware Emotion Recognition Datasets}
Most of the emotion recognition datasets in the past have either only focused on a single modality, e.g., faces or body features, or have been collected in controlled settings. For example, the GENKI database~\cite{GENKI} and the UCDSEE dataset~\cite{UCDSEE} are datasets that focus primarily on the facial expressions collected in lab settings. The Emotion Recognition in the Wild~(EmotiW) challenges~\cite{emotiw} host three databases: AFEW dataset~\cite{AFEW}~(collected from TV shows and movies), SFEW~(a subset of AFEW  with only face frames annotated), and HAPPEI database, which focuses on the problem of group-level emotion estimation. Some of the recent works have realized the potential of using context for emotion recognition and highlighted the lack of such datasets. Context-Aware Emotion Recognition~(CAER) dataset~\cite{context3} is a collection of video-clips from TV shows with $7$ discrete emotion annotations. EMOTIC dataset~\cite{context1} is a collection of images from datasets like MSCOCO~\cite{mscoco} and ADE20K~\cite{ade20k} along with images downloaded from web searches. The dataset is a collection of $23,571$ images, with about $34,320$ people annotated for $26$ discrete emotion classes.  We have summarised and compared all these datasets in Table~\ref{tab:dataanal}.
\begin{table*}[t]
\centering
\resizebox{0.8\linewidth}{!}{%
\begin{tabular}{|c|c|c|c|c|c|c|} 
\hline
\textbf{Data type } & \textbf{Dataset} & \textbf{Dataset Size} & \textbf{Agents Annotated}  & \textbf{Setting}& \textbf{Emotion Labels}& \textbf{Context}\\
\hhline{|=======|}
\multirow{5}{*}{Images }
& EMOTIC~\cite{context1}        & 18,316 images  & 34,320   & Web      & 26 Categories & Yes \\
& AffectNet~\cite{affectnet}    & 450,000 images & 450,000 & Web      & 8 Categories  & No \\
& CAER-S~\cite{context2}        & 70,000 images  & 70,000  & TV Shows & 7 Categories  & Yes \\
\hhline{|=======|}
\multirow{5}{*}{Videos}
& AFEW~\cite{AFEW}    & 1,809 clips                       & 1,809  & Movie    & 7 Categories & No\\
& CAER~\cite{context2}& 13,201 clips                      & 13,201  & TV Show   & 7 Categories & Yes \\
& IEMOCAP~\cite{iemocap}& 12 hrs                      & -  & TV Show   & 4 Categories & Yes \\
& \textbf{\dataname}  & 45 clips(10 mins each) & 3544              & Real Settings              & 4 Categories   & Yes \\
\hline
\end{tabular}
}
\caption{\textbf{Context-Aware Emotion Recognition Dataset Analysis: }We compare \dataname~with existing emotion recognition datasets such as EMOTIC~\cite{context1}, AffectNet~\cite{affectnet}, CAER and CAER-S~\cite{context2}, and AFEW~\cite{AFEW}.}
\label{tab:dataanal}
\end{table*}

\section{Our Approach: \modelname}
\label{sec:approach}
In this section, we give an overview of the approach in Section \ref{subsec:overview} and motivate the three context interpretations in Section \ref{subsec:context1}, \ref{subsec:context2}, and \ref{subsec:context3}. 
\subsection{Notation and Overview}
\label{subsec:overview}
We present an overview of our context-aware multimodal emotion recognition model, \modelname, in Figure~\ref{fig:overview}. Our input consists of an RGB image, $I$. We process $I$ to generate the input data for each network corresponding to the three contexts. The network for Context 1 consists of $n$ streams corresponding to $n$ distinct modalities denoted as $m_1, m_2, \dots, m_n$. Each distinct layer outputs a feature vector, $f_i$. The $n$ feature vectors $f_1, f_2, \ldots, f_n$ are combined via multiplicative fusion~\cite{m3er} to obtain a feature encoding, $h_1 = g(f_1, f_2, \ldots, f_n)$, where $g(\cdot)$ corresponds to the multiplicative fusion function. Similarly, $h_2$, and $h_3$ are computed through the networks corresponding to the second and third Contexts. $h_1, h_2$, and $h_3$ are concatenated to perform multi-label emotion classification. 
\subsection{Context 1: Multiple Modalities}
\label{subsec:context1}
In real life, people appear in a multi-sensory context that includes a voice, a body, and a face; these aspects are also perceived as a whole. Combining more than one modality to infer emotion is beneficial because cues from different modalities can complement each other. They also seem to perform better on in-the-wild datasets~\cite{m3er} than other unimodal approaches. Our approach is extendable to any number of modalities available. To validate this claim, other than EMOTIC and \dataname, which have two modalities, faces, and gaits, we also show results on the IEMOCAP dataset which face, text, and speech as three modalities. From the input image $I$, we obtain $m_1, m_2, \dots, m_n$ using processing steps as explained in Section \ref{subsec:data-process}. These inputs are then passed through their respective neural network architectures to obtain $f_1, f_2, \dots, f_n$.
To make our algorithm robust to sensor noise and averse to noisy signals, we combine these features multiplicatively to obtain $h_1$. As shown in previous research~\cite{multiplicative,m3er}, multiplicative fusion learns to emphasize reliable modalities and to rely less on other modalities. 
To train this, we use the modified loss function proposed previously~\cite{m3er} defined as:
\begin{equation}
    L_\textrm{multiplicative} = -\sum_{i=1}^{n} \left ( p_{i}^{e} \right )^{\frac{\beta}{n-1}} \log p_{i}^{e}
\label{eq:multloss}
\end{equation}
where $n$ is the total number of modalities being considered, and $p_{i}^{e}$ is the prediction for emotion class, $e$, given by
the network for the $i^{\textrm{th}}$ modality.
\subsection{Context 2: Situational/Background Context}
\label{subsec:context2}
Our goal is to identify semantic context from images and videos to perform perceived emotion recognition. Semantic context includes the understanding of objects --excluding the primary agent-- present in the scene, their spatial extents, keywords, and the activity being performed. For instance, in Figure~\ref{fig:cover}, the input image consists of a group of people gathered around with drinks on a bright sunny day. The ``bright sunny day'', ``drink glasses'', ``hats'' and ``green meadows'' constitute semantic components and may affect judgement of one's perceived emotion. 

Motivated by multiple approaches in the computer vision literature~\cite{semantic1,abn} surrounding semantic scene understanding, we use an attention mechanism to train a model to focus on different aspects of an image while \textit{masking} the primary agent, to extract the semantic components of the scene. The mask, $I_{\textrm{mask}}\in \mathbb{R}^{224 \times 224}$, for an input image $I$ is given as

\begin{equation}
I_{\textrm{mask}} =
     \begin{cases}
       I(i,j)  & \text{if} \ I(i,j) \not \in \textrm{bbox}_{\textrm{agent}} ,\\
      0 &\text{otherwise.}
        
     \end{cases}
     \label{eq: mask}
\end{equation}

\noindent where $\textrm{bbox}_{\textrm{agent}}$ denotes the bounding box of the agent in the scene.

\subsection{Context 3: Inter-Agent Interactions/Socio-Dynamic Context}
\label{subsec:context3}
When an agent is surrounded by other agents, their perceived emotions change. When other agents share an identity or are known to the agent, they often coordinate their behaviors. This varies when other agents are strangers. Such interactions and proximity can help us infer the emotion of agents better.

Prior experimental research has used walking speed, distance, and proximity features to model socio-dynamic interactions between agents to interpret their personality traits.
Some of these algorithms, like the social force model~\cite{socialforce}, are based on the assumption that pedestrians are subject to attractive or repulsive forces that drive their dynamics. Non-linear models like RVO~\cite{rvo} aim to model collision avoidance among individuals while walking to their individual goals. But, both of these methods do not capture cohesiveness in a group.

 We propose an approach to model these socio-dynamic interactions by computing proximity features using depth maps. The depth map, $I_{\textrm{depth}} \in \mathbb{R}^{224 \times 224}$, corresponding to input image, $I$, is represented through a 2D matrix where,

\begin{equation}
I_{\textrm{depth}} (i,j) = d(I(i,j),c) 
\label{eq: depth}
\end{equation}

\noindent $d(I(i,j),c)$ represents the distance of the pixel at the $i^{\textrm{th}}$ row and $j^{\textrm{th}}$ column from the camera center, $c$. We pass $I_{\textrm{depth}}$ as input depth maps through a CNN and obtain $h_3$. 



In addition to depth map-based representation, we also use Graph Convolutional Networks~(GCNs) to model the proximity-based socio-dynamic interactions between agents. GCNs have been used to model similar interactions in traffic networks~\cite{graph_traffic} and activity recognition~\cite{graph_activity}. The input to a GCN network consists of the spatial coordinates of all agents, denoted by $X \in \mathbb{R}^{n \times 2}$, where $n$ represents the number of agents in the image, as well as the unweighted adjacency matrix, $A\in \mathbb{R}^{n \times n}$, of the agents, which is defined as follows,

\begin{equation}
A(i,j) = 
     \begin{cases}
       e^{-d(v_i,v_j)}  & \text{if $d(v_i,v_j) < \mu$},\\
      0 &\text{otherwise.}
        
     \end{cases}
     \label{eq: similarity_function}
\end{equation}

\noindent The function $f = e^{-d(v_i,v_j)}$~\cite{belkin2003laplacian} denotes the interactions between any two agents.

\section{Network Architecture and Implementation Details}
\label{sec:network}
In this section, we elaborate on the implementation and network architectures of \modelname. The data pre-processing for the streams of \modelname~are presented in \ref{subsec:data-process}. We include details about the network architectures of context 1, context 2, and context 3 in Section~\ref{subsec:network-arch}. We explain the early fusion technique we use to fuse the features from the three context streams to infer emotion and the loss function used for training the multi-label classification problem. 
\subsection{ Data Processing}
\label{subsec:data-process}
\noindent \textbf{Context1: }We use OpenFace~\cite{openface}~to extract a $144$-dimensional face modality vector, $m_1 \in \mathbb{R}^{144}$ obtained through multiple facial landmarks. We compute the 2D gait modality vectors, $m_2 \in \mathbb{R}^{25 \times 2}$ using OpenPose~\cite{openpose}~to extract $25$-coordinates from the input image $I$. For each coordinate, we record the $x$ and $y$ pixel values.\\\\
\noindent \textbf{Context2: }We use RobustTP~\cite{robusttp}, which is a pedestrian tracking method to compute the bounding boxes for all agents in the scene. These bounding boxes are used to compute $I_{\textrm{mask}}$ according to Equation~\ref{eq: mask}.\\\\
\noindent \textbf{Context3: }We use Megadepth~\cite{megadepth} to extract the depth maps from the input image $I$. The depth map, $I_{\textrm{depth}}$, is computed using Equation~\ref{eq: depth}. 
\subsection{Network Architecture}
\label{subsec:network-arch}
\noindent \textbf{Context1: }Given a face vector, $m_1$, we use three $1$D convolutions~(depicted in light green color in Figure~\ref{fig:overview}) with batch normalization and ReLU non-linearity. This is followed by a max pool operation and three fully-connected layers~(cyan color in Figure~\ref{fig:overview}) with batch normalization and ReLU. For $m_2$, we use the ST-GCN architecture proposed by \cite{step}, which is currently the SOTA network for emotion classification using gaits. Their method was originally designed to deal with 2D pose information for 16 body joints. We modify their setup for 2D pose inputs for $25$ joints. We show the different layers and hyper-parameters used in Figure \ref{fig:overview}. The two networks give us $f_1$ and $f_2$, which are then multiplicatively fused~(depicted in red color in Figure~\ref{fig:overview}) to generate $h_1$.\\\\
\noindent \textbf{Context 2: }For learning the semantic context of the input image $I$, we use the Attention Branch Network (ABN) \cite{abn} on the masked image $I_{\textrm{mask}}$. ABN contains an attention branch which focuses on attention maps to recognize and localize important regions in an image. It outputs these potentially important locations in the form of $h_2$.\\\\
\noindent \textbf{Context 3: } We perform two experiments using both depth map and a GCN. For depth-based network, we compute the depth map, $I_\textrm{depth}$ and pass it through a CNN. The CNN is composed of $5$ alternating $2$D convolutional layers~(depicted in dark green color in Figure~\ref{fig:overview}) and max pooling layers~(magenta color in Figure~\ref{fig:overview}). This is followed by two fully connected layers of dimensions $1000$ and $26$~(cyan color in Figure~\ref{fig:overview}).

For the graph-based network, we use two graph convolutional layers followed by two linear layers of dimension $100$ and $26$. \\\\
\noindent \textbf{Fusing Context Interpretations: }To fuse the feature vectors from the three context interpretations, we use an early fusion technique. We concatenate the feature vectors before making any individual emotion inferences. 
\[h_{\textrm{concat}} = \left[h_1, h_2, h_3 \right]\]
We use two fully connected layers of dimensions $52$ and $26$, followed by a softmax layer. This output is used for computing the loss and the error, and then back-propagating the error back to the network.\\\\
\noindent \textbf{Loss Function: } Our classification problem is a multi-label classification problem where we assign one or more than one emotion label to an input image or video. To train this network, we use the multi-label soft margin loss function and denote it by $L_\textrm{classification}$. The loss function optimizes a multi-label one-versus-all loss based on max-entropy between the input $x$ and output $y$.

So, we combine the two loss functions, $L_\textrm{multiplicative}$~(from Eq.~\ref{eq:multloss}) and $L_\textrm{classification}$ to train \modelname.
\begin{equation}
L_\textrm{total} = \lambda_1 L_\textrm{multiplicative} + \lambda_2 L_\textrm{classification} 
\end{equation}

\section{Datasets}
\label{sec:dataset}
In Section~\ref{subsec:old}, we give details about the benchmark dataset for context-aware emotion recognition, EMOTIC. We present details about the new dataset, \dataname~and also perform a comparison with other existing datasets in Section~\ref{subsec:new}. Like summarised in Table~\ref{tab:dataanal}, there are a lot more datasets for emotion recognition, but they do not have any context available. Though our approach will work on these datasets, we do not expect any significant improvement over the SOTA on these datasets. Just to reinforce this, we did run our method on IEMOCAP~\cite{iemocap}, which has limited context information, and summarise our results in Appendix~\ref{appendix:2}.
\subsection{EMOTIC Dataset}
\label{subsec:old}
The EMOTIC dataset contains 23,571 images of 34,320 annotated people in unconstrained environments. The annotations consist of the apparent emotional states of the people in the images. Each person is annotated for 26 discrete categories, with multiple labels assigned to each image. 
\subsection{\dataname~Dataset}
\label{subsec:new}
\subsubsection{Annotation}
\dataname~consists of 45 videos that were captured using stationary cameras in $8$ real-world setting including a hospital entrance, an institutional building, a bus stop, a train station, and a marketplace, a tourist attraction, a shopping place and more. The annotators annotated agents with clearly visible faces and gaits across all videos. $10$ annotators annotated a total of $3544$ agents. The annotations consist of the following emotion labels-- Angry, Happy, Neutral, and Sad. Efforts to build on this dataset are still ongoing. The dataset collected and annotated so far can be found at the Project webpage. To prepare train and test splits for the dataset, we randomly selected $36$ videos for the training and $9$ videos for testing. 

While perceived emotions are essential, other affects such as dominance and friendliness are important for carrying out joint and/or group tasks. Thus, we additionally label each agent for dominance and friendliness. More details about the annotation process, labelers and labels processing are presented in Appendix~\ref{appendix:1}.


\section{Experiments and Results}
In this section, we discuss the experiments conducted for \modelname. We present details on hyperparameters and training details in Section~\ref{subsec:training}. In section~\ref{subsec:metric}, we list the prior methods we compare the performance of \modelname~with. We present an elaborate analysis of both qualitative and quantitative results in Section~\ref{subsec:analysis}. In Section~\ref{subsec:ablation}, we perform experiments to validate the importance of each component of \modelname.
\begin{figure*}[t]
    \centering
    \includegraphics[width=\textwidth]{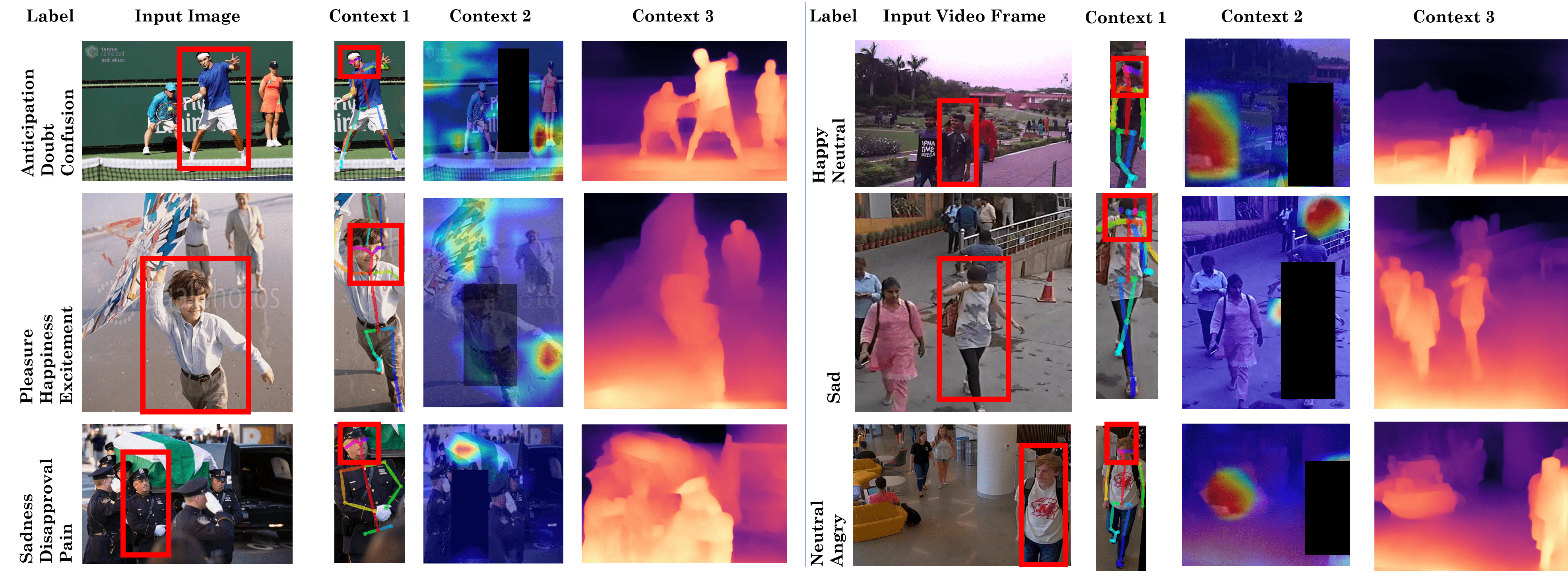}
    \caption{\textbf{Qualitative Results: }We show the classification results on three examples, each from the EMOTIC dataset~(left) and \dataname~Dataset~(right), respectively. In the top row example (left) and middle row example (right), the depth map clearly marks the tennis player about to swing to convey anticipation, and the woman coming from the hospital to convey sadness, respectively. In the bottom row (left) and bottom row (middle) examples, the semantic context of the coffin and the child's kite is clearly identified to convey sadness and pleasure, respectively.}
    \label{fig:results}
\end{figure*}
\subsection{Training Details}
\label{subsec:training}
For training \modelname~on the EMOTIC dataset, we use the standard train, val, and test split ratios provided in the dataset. For \dataname, we split the dataset into 85\% training (85\%) and testing (15\%) sets. In \dataname~each sample point is an agent ID; hence the input is all the frames for the agent in the video. To extend \modelname~on videos, we perform a forward pass for all the frames and take an average of the prediction vector across all the frames and then compute the AP scores and use this for loss calculation and backpropagating the loss.  We use a batch size of $32$ for EMOTIC and a batchsize of $1$ for \dataname. We train \modelname~for $75$ epochs. We use the Adam optimizer with a learning rate of $0.0001$. All our results were generated on NVIDIA GeForce GTX 1080 Ti GPU. All the code was implemented using PyTorch~\cite{pytorch}.
\subsection{Evaluation Metrics and Methods}
\label{subsec:metric}
\noindent We use the standard metric Average Precision (AP) to evaluate all our methods. For both EMOTIC and \dataname~datasets, we compare our methods with the following SOTA methods. 
\begin{enumerate}[noitemsep]
    \item \textbf{Kosti et al.~\cite{context1} }propose a two-stream network followed by a fusion network. The first stream encodes context and then feeds the entire image as an input to the CNN. The second stream is a CNN for extracting body features. The fusion network combines features of the two CNNs and estimates the discrete emotion categories. 
    \item \textbf{Zhang et al.~\cite{context3} }build an affective graph with nodes as the context elements extracted from the image. To detect the context elements, they use a Region Proposal Network~(RPN). This graph is fed into a Graph Convolutional Network~(GCN). Another parallel branch in the network encodes the body features using a CNN. The outputs from both the branches are concatenated to infer an emotion label.
    \item \textbf{Lee et al.~\cite{context2} }present a network architecture, CAER-Net consisting of two subnetworks, a two-stream encoding network, and an adaptive fusion network. The two-stream encoding network consists of a face stream and a context-stream where facial expression and context~(background) are encoded. An adaptive fusion network is used to fuse the two streams. 
\end{enumerate}
We use the publicly available implementation for Kosti et al.~\cite{context1} and train the entire model on \dataname. Both Zhang et al.~\cite{context3} and Lee et al.~\cite{context2} do not have publicly available implementations. We reproduce the method by Lee et al.~\cite{context2} to the best of our understanding. For Zhang et al.~\cite{context3}, while we report their performance on the EMOTIC dataset, with limited implementation details, it was difficult to build their model to test their performance on \dataname.
\begin{table}[h!]
\centering
(a) AP Scores for EMOTIC Dataset.
\resizebox{\columnwidth}{!}{%
\begin{tabular}{|l|c|c|c|c|c|}
\hline
\textbf{Labels}          & Kosti et al.\cite{context1} & Zhang et al.\cite{context3} & Lee et al.\cite{context2} & \multicolumn{2}{c|}{\textbf{\modelname}} \\
&  &  &  & GCN-Based & Depth-Based \\ \hline
Affection       & 27.85   & 46.89   & 19.9    & 36.78 & \textbf{45.23}\\ \hline
Anger           & 09.49   & 10.87   & 11.5    & 14.92 & \textbf{15.46}\\ \hline
Annoyance       & 14.06   & 11.23   & 16.4    & 18.45 & \textbf{21.92}\\ \hline
Anticipation    & 58.64   & 62.64   & 53.05   & 68.12 & \textbf{72.12}\\ \hline
Aversion        & 07.48   & 5.93    & 16.2    & 16.48 & \textbf{17.81} \\ \hline
Confidence      & 78.35   & 72.49   & 32.34   & 59.23 & \textbf{68.65} \\ \hline
Disapproval     & 14.97   & 11.28   & 16.04   & \textbf{21.21} & 19.82 \\ \hline
Disconnection   & 21.32   & 26.91   & 22.80   & 25.17 & \textbf{43.12} \\ \hline
Disquietment    & 16.89   & 16.94   & 17.19   & 16.41 & \textbf{18.73} \\ \hline
Doubt/Confusion & 29.63   & 18.68   & 28.98   & 33.15 & \textbf{35.12}\\ \hline
Embarrassment   & 03.18   & 1.94    & 15.68   & 11.25 & \textbf{14.37} \\ \hline
Engagement      & 87.53   & 88.56   & 46.58   & 90.45 & \textbf{91.12}\\ \hline
Esteem          & 17.73   & 13.33   & 19.26   & 22.23 & \textbf{23.62} \\ \hline
Excitement      & 77.16   & 71.89   & 35.26   & 82.21 & \textbf{83.26} \\ \hline
Fatigue         & 09.70   & 13.26   & 13.04   & \textbf{19.15} & 16.23 \\ \hline
Fear            & 14.14   & 4.21    & 10.41   & 11.32 & \textbf{23.65} \\ \hline
Happiness       & 58.26   & 73.26   & 49.36   & 68.21 & \textbf{74.71} \\ \hline
Pain            & 08.94   & 6.52    & 10.36   & 12.54 & \textbf{13.21} \\ \hline
Peace           & 21.56   & 32.85   & 16.72   & \textbf{35.14} & 34.27 \\ \hline
Pleasure        & 45.46   & 57.46   & 19.47   & 61.34 & \textbf{65.53} \\ \hline
Sadness         & 19.66   & 25.42   & 11.45   & \textbf{26.15} & 23.41 \\ \hline
Sensitivity     & 09.28   & 5.99    & \textbf{10.34}   &  9.21 & 8.32 \\ \hline
Suffering       & 18.84   & 23.39   & 11.68   & 22.81 & \textbf{26.39}\\ \hline
Surprise        & \textbf{18.81}   & 9.02    & 10.92   & 14.21 & 17.37 \\ \hline
Sympathy        & 14.71   & 17.53   & 17.125  & 24.63 & \textbf{34.28} \\ \hline
Yearning        & 08.34   & 10.55   & 9.79    & 12.23 & \textbf{14.29} \\ \hline\hline
\textbf{mAP}   & 27.38   & 28.42   & 20.84   & 32.03 & \textbf{35.48} \\\hline
\end{tabular}
}

\bigskip 
\centering
(b) AP Scores for \dataname~Dataset.
\centering
\resizebox{\columnwidth}{!}{%
\begin{tabular}{|l|c|c|c|c|c|c|}
\hline
\textbf{Labels}          & Kosti et al.\cite{context1} & Zhang et al.\cite{context3} & Lee et al.\cite{context2} & \multicolumn{2}{c|}{\textbf{\modelname}} \\
&  &  &  & GCN-Based & Depth-Based \\ \hline
Anger & 58.46 & - & 42.31 & 65.13 & \textbf{69.42} \\ \hline
Happy & 69.12 & - & 56.79 & 72.46 & \textbf{73.18} \\ \hline
Neutral & 42.27 & - & 39.24 & 44.51 & \textbf{48.51} \\ \hline
Sad & 63.83 & - & 54.33 & 68.25 & \textbf{72.24} \\ \hline\hline
\textbf{mAP} & 58.42 & - & 48.21 & 62.58 & \textbf{65.83} \\
\hline
\end{tabular}
}
\caption{\small{\textbf{Emotion Classification Performance: }We report the AP scores on the EMOTIC and the \dataname~datasets. \modelname~outperforms all the three methods for most of the classes and also overall.}}
\label{tab:accuracy}
\vspace{-15pt}
\end{table}

\subsection{Analysis and Discussion}
\label{subsec:analysis}

\noindent \textbf{Comparison with SOTA:} We summarize the evaluation of the APs for all the methods on the EMOTIC and \dataname~datasets in Table~\ref{tab:accuracy}. For \modelname, we report the AP scores for both GCN-based and Depth Map-based implementations of Context 3. On both the EMOTIC and \dataname~datasets, \modelname~outperforms the SOTA.

\noindent \textbf{Generalize to more Modalities:} A major factor for the success of \modelname~is its ability to combine different modalities effectively via multiplicative fusion. Our approach learns to assign higher weights to more expressive modalities while suppressing weaker ones. For example, in instances where the face may not be visible,~\modelname infers the emotion from context (See Figure~\ref{fig:results}, middle row(right)). This is in contrast to Lee et al.~\cite{context2}, which relies on the availability of face data. Consequently, they perform poorly on both the EMOTIC and \dataname~datasets, as both datasets contain many examples where the face is not visible clearly. To further demonstrate the ability of \modelname~to generalize to any modality, we additionally report our performance on the IEMOCAP dataset~\cite{iemocap} in Appendix~\ref{appendix:2}.

\noindent \textbf{GCN versus Depth Maps:} GCN-based methods do not perform as well as depth-based but are a close second. This may be due to the fact that, on average most images of the EMOTIC dataset contain $5$ agents. GCN-based methods in the literature have been trained on datasets with a lot more number of agents in each image or video. Moreover, with a depth-based approach, \modelname~leans a $3$D aspect of the scene in general and is not limited to inter-agent interactions. 

\begin{figure}[t]
    \centering
    \includegraphics[width=0.8\columnwidth]{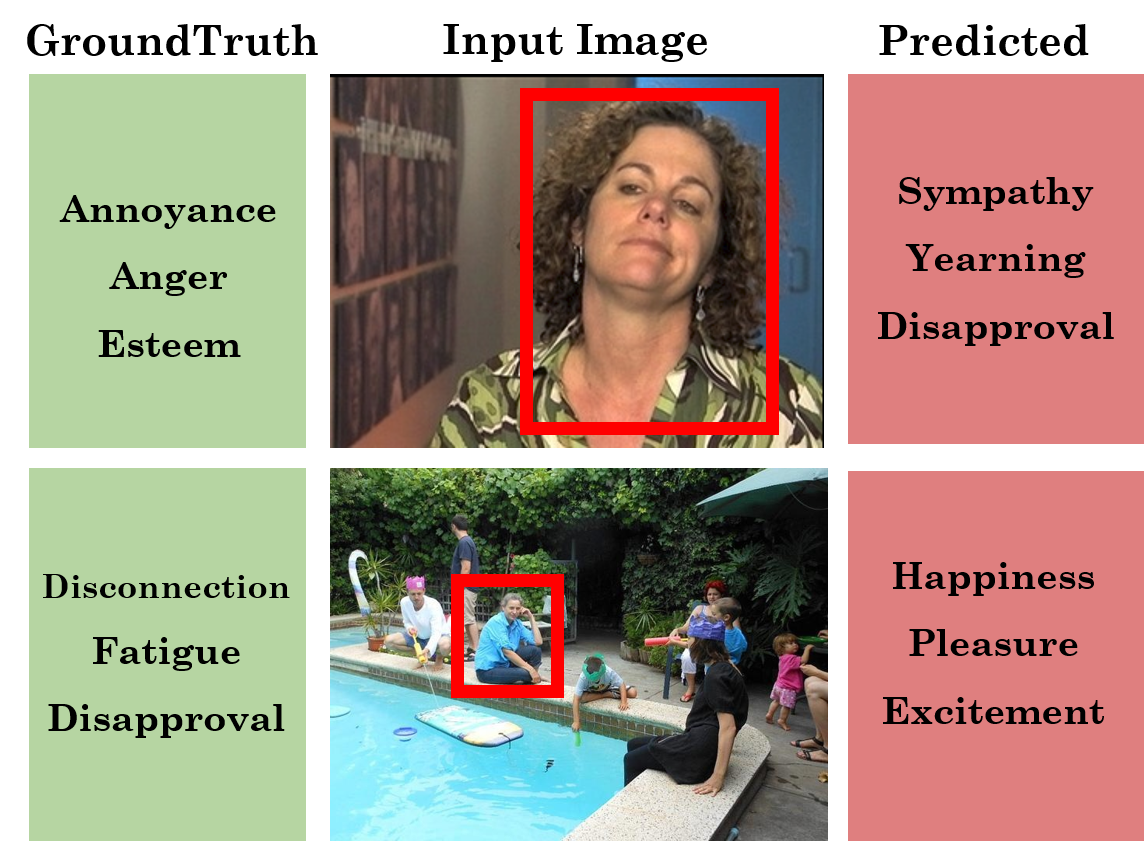}
    \caption{\small{\textbf{Misclassification by \modelname:} We show two examples where \modelname~incorrectly classifies the labels. In the first examples, \modelname~is confused about the prediction due to lack of any context. In the second example, there is a lot of context available, which also becomes confusing. }
    }
    \label{fig:failure}
\vspace{-10pt}
\end{figure}

\noindent\textbf{Failure Cases: }We show two examples from EMOTIC dataset in Figure~\ref{fig:failure} where \modelname~fails to classify correctly. We also show the ground-truth and predicted emotion labels. In the first image, \modelname is unable to gather any context information. On the other hand, in the second image, there is a lot of context information like the many visual elements in the image and multiple agents. This leads to an incorrect inference of the perceived emotion. 

\begin{table}[h!]
\centering
(a) Ablation Experiments on EMOTIC Dataset.
\resizebox{0.85\columnwidth}{!}{%
\begin{tabular}{|l|c|c|c|c|}
\hline
\multirow{2}{*}{\textbf{Labels}} & \multicolumn{4}{c|}{Context Interpretations}\\\cline{2-5}
& Only 1 & Only 1 and 2 & Only 1 and 3 & {\textbf{1, 2 and 3}} \\ \hline 
Affection       & 29.87   & 41.83 & 30.15 & 45.23 \\\hline
Anger           & 08.52   & 11.41 & 8.36 & 15.46 \\ \hline
Annoyance       & 09.65   & 17.37 & 12.91 & 21.92 \\ \hline
Anticipation    & 46.23   & 67.59 & 60.53 & 72.12\\ \hline
Aversion        & 06.27   & 11.71 & 09.46 & 17.81 \\ \hline
Confidence      & 51.92   & 65.27 & 59.63 & 68.85 \\ \hline
Disapproval     & 11.81   & 17.35 & 15.41 & 19.82 \\ \hline
Disconnection   & 31.74   & 41.46 & 32.56 & 43.12 \\ \hline
Disquietment    & 07.57   & 12.69 & 12.24 & 18.73 \\ \hline
Doubt/Confusion & 21.62   & 31.28 & 29.51 & 35.12 \\ \hline
Embarrassment   & 08.43   & 10.51 & 12.25 & 14.37 \\ \hline
Engagement      & 78.68   & 84.62 & 81.51 & 91.12 \\ \hline
Esteem          & 18.32   & 18.79 & 09.42 & 23.62 \\ \hline
Excitement      & 73.19   & 80.54 & 76.14 & 83.26 \\ \hline
Fatigue         & 06.34   & 11.95 & 14.15 & 16.23 \\ \hline
Fear            & 14.29   & 21.36 & 22.29 & 23.65 \\ \hline
Happiness       & 52.52   & 69.51 & 71.51 & 74.71 \\ \hline
Pain            & 05.75   & 09.56 & 11.10 & 13.21 \\ \hline
Peace           & 13.53   & 30.72 & 30.15 & 34.27 \\ \hline
Pleasure        & 58.26   & 61.89 & 59.81 & 65.53 \\ \hline
Sadness         & 19.94   & 19.74 & 22.27 & 23.41 \\ \hline
Sensitivity     & 03.16   & 04.11 & 8.15 & 8.32 \\ \hline
Suffering       & 15.38   & 20.92 & 12.83 & 26.39 \\ \hline
Surprise        & 05.29   & 16.45 & 16.26 & 17.37 \\ \hline
Sympathy        & 22.38   & 30.68 & 22.17 &  34.28\\ \hline
Yearning        & 04.94   & 10.53 & 9.82 & 14.29\\ \hline\hline
\textbf{mAP}    & 24.06   & 31.53 & 29.63 & 35.48 \\\hline
\end{tabular}
}

\bigskip 

\centering

(b) Ablation Experiments on \dataname~Dataset.
\centering
\resizebox{0.8\columnwidth}{!}{%
\begin{tabular}{|l|c|c|c|c|}
\hline
\multirow{2}{*}{\textbf{Labels}} & \multicolumn{4}{c|}{Context Interpretations}\\\cline{2-5}
& Only 1 & Only 1 and 2 & Only 1 and 3 & {\textbf{1, 2 and 3}} \\ \hline 
Anger & 58.51 & 63.83 & 66.15 & 69.42 \\ \hline
Happy & 61.24 & 64.16 & 68.87 & 73.18 \\ \hline
Neutral & 40.36 & 41.57 & 44.15 & 48.51 \\ \hline
Sad & 62.17 & 67.22 & 70.35 & 72.24 \\ \hline\hline
\textbf{mAP} & 55.57 & 59.20 & 62.38 & 65.83 \\
\hline
\end{tabular}
}
\caption{\small{\textbf{Ablation Experiments: }Keeping the Context interpretation 1 throughout, we remove the other two Context interpretations one by one and compare the AP scores for emotion classification on both the datasets. }}
\label{tab:ablation}
\vspace{-15pt}
\end{table}

\subsection{Qualitative Results}
We show qualitative results for three examples, each from both the datasets, respectively, in Figure~\ref{fig:results}. The first column is the input image marking the primary agents, the second column shows the corresponding extracted face and gait, the third column shows the attention maps learned by the model, and lastly, in the fourth column, we show the depth map extracted from the input image. 

The heatmaps in the attention maps indicate what the network has learned. In the bottom row (left) and bottom row (middle) examples, the semantic context of the coffin and the child's kite is clearly identified to convey sadness and pleasure, respectively. The depth maps corresponding to the input images capture the idea of proximity and inter-agent interactions. In the top row example (left) and middle row example (right), the depth map clearly marks the tennis player about to swing to convey anticipation, and the woman coming from the hospital to convey sadness, respectively.

\subsection{Ablation Experiments}
\label{subsec:ablation}
To motivate the importance of Context 2 and Context 3, we run \modelname~on both EMOTIC and \dataname~dataset removing the networks corresponding to both contexts, followed by removing either of them one by one. The results of the ablation experiments have been summarized in Table~\ref{tab:ablation}. We choose to retain Context 1 in all these runs because it is only Context 1 that is capturing information from the agent itself. 

We observe from the qualitative results in Figure~\ref{fig:results} that Context 2 seems more expressive in the images of EMOTIC dataset, while Context 3 is more representative in \dataname. This is supported by the results reported in Table~\ref{tab:ablation}, columns 2 and 3. To understand why this happens, we analyse the two datasets closely. EMOTIC dataset was collected for the task of emotion recognition with context. it is a dataset of pictures collected from multiple datasets and scraped from the Internet. As a result, most of these images have a rich background context. Moreover we also found that more than half the images of EMOTIC contain at most $3$ people. These are the reasons we believe that interpretation 2 helps more in EMOTIC than interpretation 3. In the GroupWalk Dataset, the opposite is true. The number of people per frame is much higher. This density gets captured best in interpretation 3 helping the network to make the better inference.

\section{Conclusion, Limitations, and Future Work}
We present~\modelname, a context-aware emotion recognition system that borrows and incorporates the context interpretations from psychology. We use multiple modalities~(faces and gaits), situational context, and also the socio-dynamic context information. We make an effort to use easily available modalities that can be easily captured or extracted using commodity hardware~(e.g., cameras). To foster more research on emotion recognition with naturalistic modalities, we also release a new dataset called \dataname. Our model has limitations and often confuses between certain class labels. Further, we currently
perform multi-class classification over discrete emotion labels. In the future, we would also like to move towards the continuous model of emotions~(Valence, Arousal, and Dominance). As part of future work, we would also explore more such context interpretations to improve the accuracies.

\section{Acknowledgements}
This research was supported in part by ARO Grants W911NF1910069 and W911NF1910315 and Intel.

{\small
\bibliographystyle{ieee_fullname}
\bibliography{egbib}
}
\newpage
\appendix
\section{\dataname}
\label{appendix:1}
\subsection{Annotation Procedure}
We present the human annotated GroupWalk data set which consists of 45 videos captured using stationary cameras in 8 real-world setting including a hospital entrance, an institutional building, a bus stop, a train station, and a marketplace, a tourist attraction, a shopping place and more. 10 annotators annotated 3544 agents with clearly visible faces and gaits across all videos. They were allowed to view the videos as many times as they wanted and had to categorise the emotion they perceived looking at the agent into 7 categories - "Somewhat Happy", "Extremely Happy", "Somewhat Sad", Extremely Sad", "Somewhat Angry", "Extremely Angry", "Neutral". In addition to perceived emotions, the annotators were also asked to annotate the agents in terms of dominance (5 categories- "Somewhat Submissive", "Extremely Submissive", "Somewhat Dominant", "Extremely Dominant", "Neutral" ) and friendliness (5 categories- "Somewhat Friendly", "Extremely Friendly", "Somewhat Unfriendly", "Extremely Unfriendly", "Neutral"). Attempts to build the dataset are still ongoing. 

For the sake of completeness, we show the friendliness label distribution and dominance label distribution for every annotator in Figure~\ref{three} and Figure~\ref{four} respectively.
\label{section_annotproc}
\subsection{Labels Processing}

4 major labels that have been considered are Angry, Happy, Neutral and Sad. As described in Section \ref{section_annotproc}, one can observe that the annotations are either "Extreme" or "Somewhat" variants of these major labels (except Neutral). Target labels were now generated for each agent. Each of them are of the size $1$ x $4$ with the $4$ columns representing the $4$ emotions being considered and are initially all $0$. For a particular agent id, if the annotation by an annotator was an "Extreme" variant of Happy, Sad or Angry, $2$ was added to the number in the column representing the corresponding major label. Otherwise for all the other cases, $1$ was added to the number in the column representing the corresponding major label. Once we have gone through the entire dataset, we normalize the target label vector so that vector is a combination of only $1$s and $0$s.      
\begin{figure}[]
\centering
\includegraphics[width=.5\textwidth]{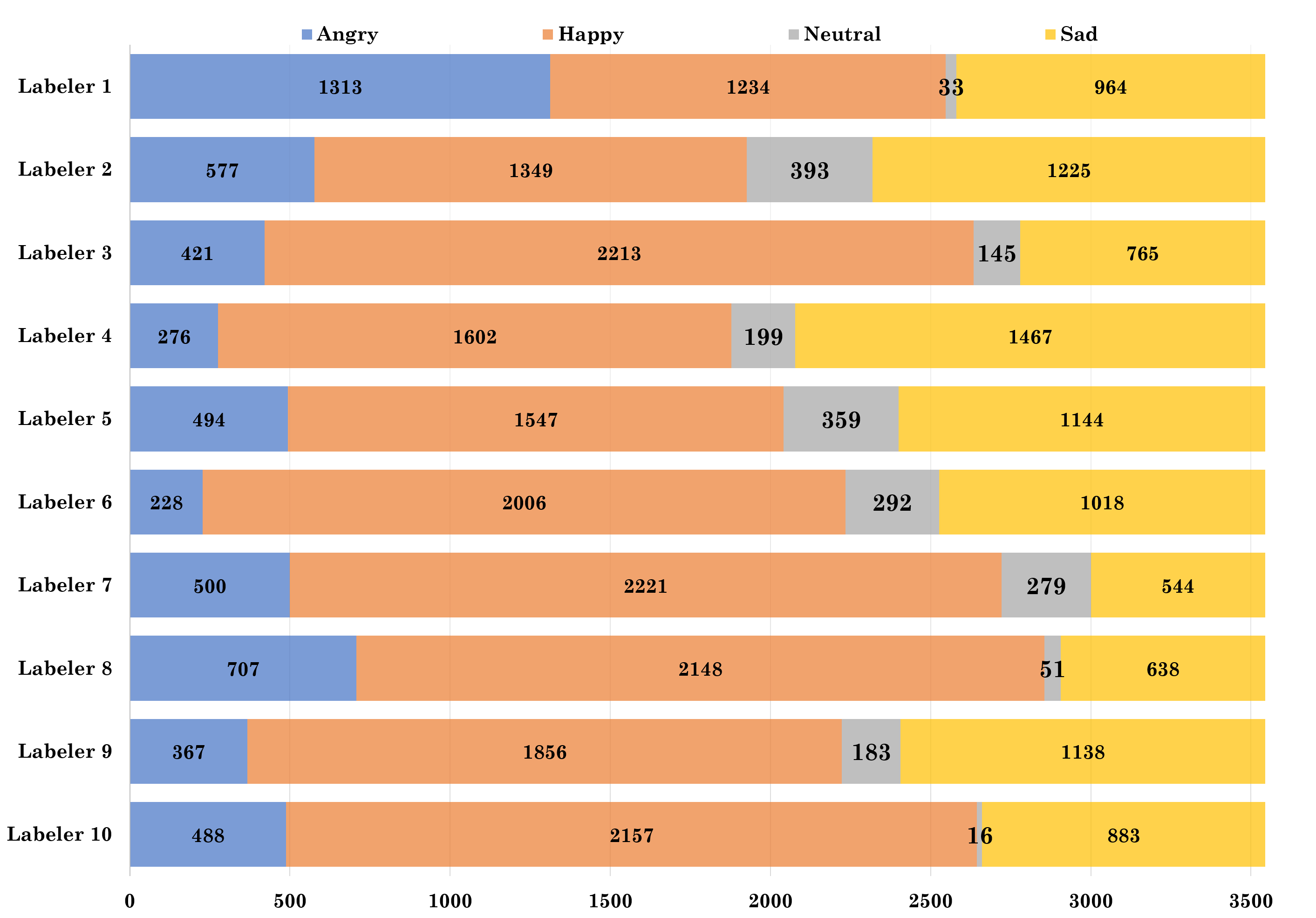}
\caption{\small{\textbf{Annotator Annotations of \dataname Dataset: }We depict the emotion class labels for \dataname~by 10 annotators. A total of $3544$ agents were annotated from 45 videos.}}
\label{one}
\bigbreak
\includegraphics[width=.5\textwidth]{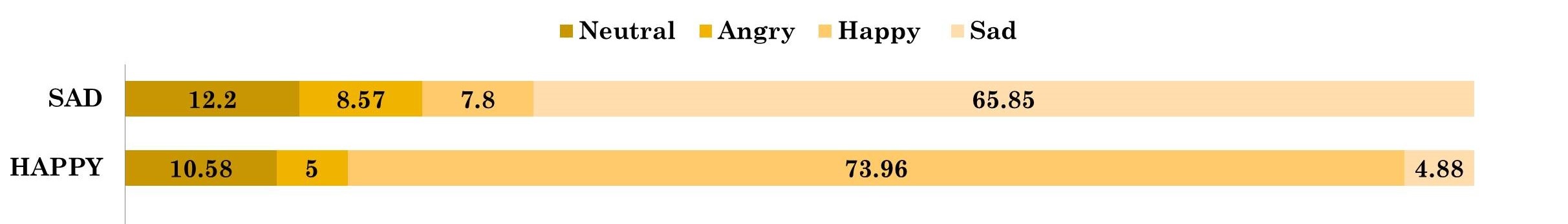}
\caption{\small{\textbf{Annotator Agreement/Disagreement: }For two emotion classes~(Happy and Sad), we depict the trend of annotator disagreement.}}
\label{two}
\end{figure}
\subsection{Analysis}
\begin{figure}[h]
\centering
\includegraphics[width=.5\textwidth]{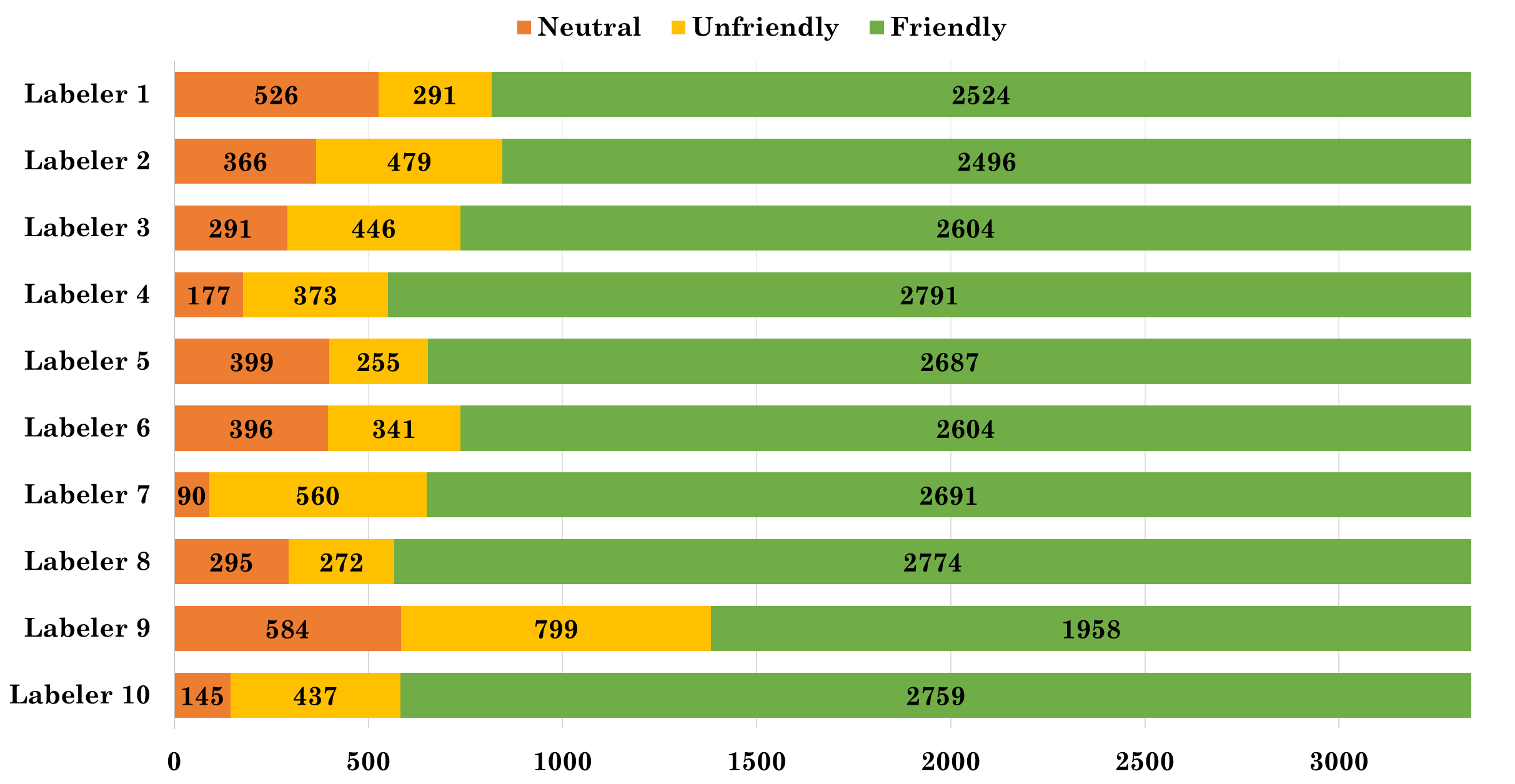}
\caption{\small{\textbf{Friendliness Labeler Annotations: }We depict the friendliness labels for \dataname~by 10 labelers. A total of $3341$ agents were annotated from 45 videos.}}
\label{three}
\bigbreak
\includegraphics[width=.5\textwidth]{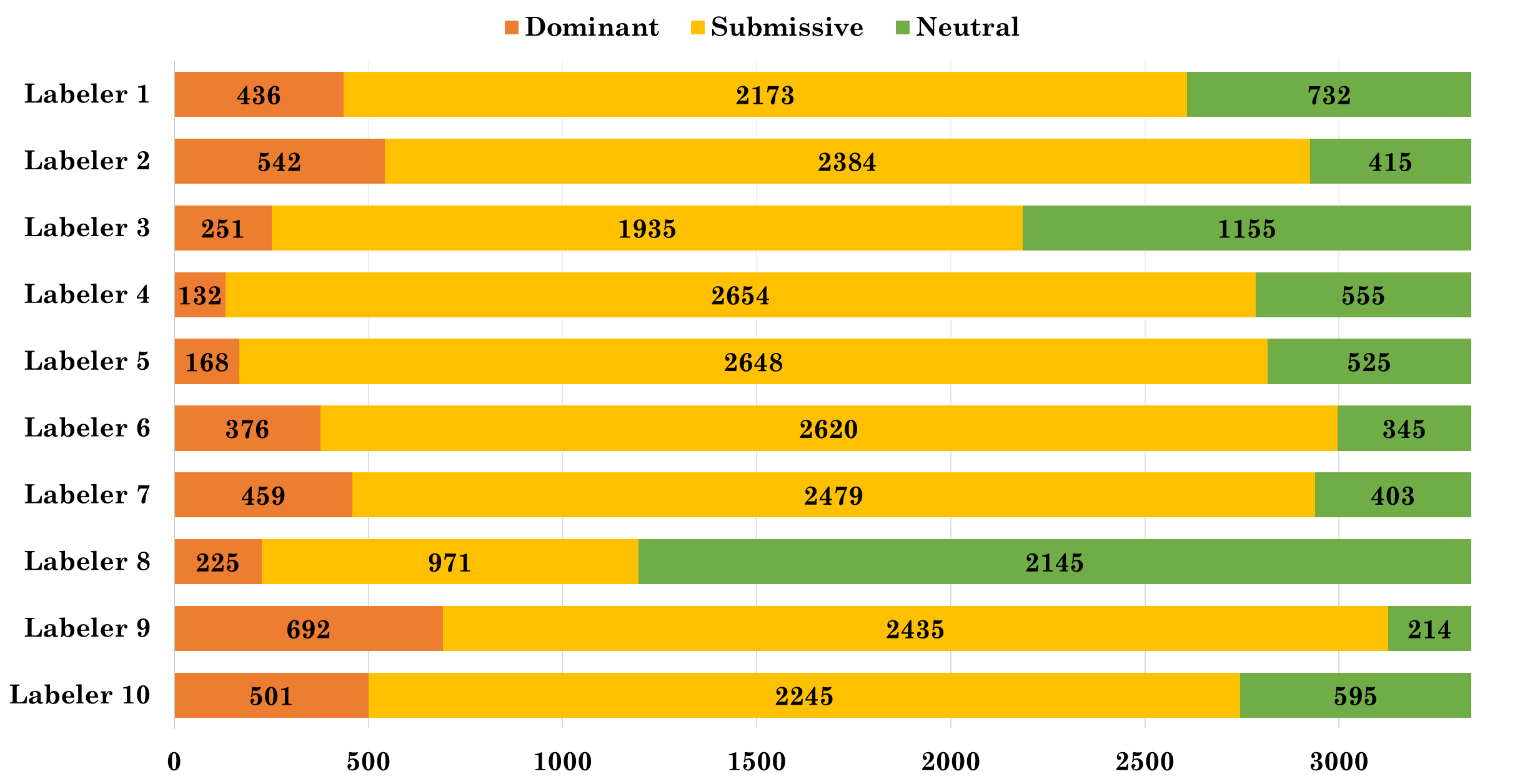}
\caption{\small{\textbf{Dominance Labeler Annotations: }We depict the dominance labels for \dataname~by 10 labelers. A total of $3341$ agents were annotated from 45 videos.}}
\label{four}
\end{figure}

We show the emotion label distribution for every annotator in Figure~\ref{one}. To understand the trend of annotator agreement and disagreement across the $10$ annotators, we gather agents labeled similarly in majority~(more than 50\% of annotators annotated the agent with the same labels) and then study the classes they were confused most with. We show this pictorially for two classes Happy and Sad in Figure~\ref{two}. For instance, we see that Happy and Sad labels are often confused with label Neutral. In addition, we also show the label distributions for every annotator for Friendliness as well as Dominance in Figure~\ref{three} and Figure~\ref{four} respectively.
%
\section{\modelname~on IEMOCAP Dataset}
\label{appendix:2}
To validate that \modelname~can be generalised for any number of modalities, we report our performance on IEMOCAP~\cite{iemocap} in Table \ref{iemocap}. IEMOCAP dataset consists of speech, text and face modalities of 10 actors recorded in the form of conversations~(both spontaneous and scripted) using a Motion Capture Camera. The labeled annotations consist of $4$ emotions -- angry, happy, neutral, and sad. This is a single-label classification as opposed to multi-label classification we reported for EMOTIC and \dataname. Because of this we choose to report mean classification accuracies rather than AP scores. Most prior work which have shown results on IEMOCAP dataset, report mean classification accuracies too. 
\begin{table}[h!]
\centering
\resizebox{\columnwidth}{!}{%
\begin{tabular}{|l|c|c|c|c|c|c|}
\hline
\textbf{Labels}          & Kosti et al.\cite{context1} & Zhang et al.\cite{context3} & Lee et al.\cite{context2} & \multicolumn{2}{c|}{\textbf{\modelname}} \\
&  &  &  & GCN-Based & Depth-Based \\ \hline
Anger & 80.7\% & -  & 77.3\% & 87.2\% & \textbf{88.2\%} \\ \hline
Happy & 78.9\%& - & 72.4\% & 82.4\% & \textbf{83.4\%} \\ \hline
Neutral & 73.5\% & - & 62.8\% & 75.5\% & \textbf{77.5\%} \\ \hline
Sad & 81.3\% & - & 68.7\% & 88.2\% & \textbf{88.9\%} \\ \hline\hline
\textbf{mAP} & 78.6\% & - & 70.3\% & 83.4\%& \textbf{84.5\%} \\
\hline
\end{tabular}
}
\caption{\small{\textbf{IEMOCAP Experiments: }Mean Classification Accuracies for IEMOCAP Dataset.}}
\label{iemocap}
\end{table}

As can be seen from the Table \ref{iemocap}, there is not a significant improvement in the accuracy, $84.5\%$ as SOTA works, not essentially based on context have reported an accuracy of $82.7\%$. We believe that the controlled settings in which the dataset is collected, with minimal context information results in not huge improvements. Moreover we also see that prior works in context, Kosti et al.~\cite{context1} and Lee et al.~\cite{context3} sort of do not get any context to learn from and hence do not perform so well. Even \modelname's performance is a result of incorporating modalities, with small contribution from context.

 \newpage

\end{document}